\documentclass[10pt,a4paper]{article}
\usepackage{lrec-coling2024}
\usepackage{tikz}
\usetikzlibrary{shapes, backgrounds}
\usepackage{graphicx}
\usepackage{booktabs}
\usepackage{multicol}
\usepackage{multirow}
\usepackage{inconsolata}
\usepackage{mathtools}
\usepackage{tabularx}
\usepackage{xspace}
\newcommand{\corpusname}{\textsc{DeFaBel}\xspace}
\usepackage{paralist}

\makeatletter
\renewcommand\paragraph{\@startsection{paragraph}{4}{\z@}%
  {0.8ex \@plus1ex \@minus.2ex}%
  {-1em}%
  {\normalfont\normalsize\bfseries}}
\makeatother

\title{Can Factual Statements be Deceptive? \\ The \corpusname Corpus of Belief-based Deception}

\name{Aswathy Velutharambath$^{1,2}$, Amelie Wührl$^{1}$, and Roman Klinger$^{1,3}$} 

\address{%
  $^{1}$Institut f\"ur Maschinelle Sprachverarbeitung, University of Stuttgart, Germany\\
   $^{2}$Psychological AI (100 Worte Sprachanalyse GmbH), Heilbronn, Germany \\
   $^{3}$Fundamentals of Natural Language Processing, University of Bamberg, Germany \\
   \texttt{aswathy.velutharambath@100worte.de,
     amelie.wuehrl@ims.uni-stuttgart.de}\\
      \texttt{roman.klinger@uni-bamberg.de}\\
}

\abstract{If a person firmly believes in a non-factual statement, such
  as ``\textit{The Earth is flat}'', and argues in its favor, there is
  no inherent intention to deceive. As the argumentation stems from
  genuine belief, it may be unlikely to exhibit the linguistic
  properties associated with deception or lying. This interplay of
  factuality, personal belief, and intent to deceive remains an
  understudied area. Disentangling the influence of these variables in
  argumentation is crucial to gain a better understanding of the
  linguistic properties attributed to each of them. To study the
  relation between deception and factuality, based on belief, we
  present the \textsc{DeFaBel} corpus, a crowd-sourced resource of
  belief-based deception. To create this corpus, we devise a study in
  which participants are instructed to write arguments supporting
  statements like ``\textit{eating watermelon seeds can cause
    indigestion}'', regardless of its factual accuracy or their
  personal beliefs about the statement. In addition to the generation
  task, we ask them to disclose their belief about the statement. The
  collected instances are labelled as deceptive if the arguments are
  in contradiction to the participants' personal beliefs. Each
  instance in the corpus is thus annotated (or implicitly labelled)
  with personal beliefs of the author, factuality of the statement,
  and the intended deceptiveness. The \textsc{DeFaBel} corpus contains
  1031 texts in German, out of which 643 are deceptive and 388 are
  non-deceptive. It is the first publicly available corpus for
  studying deception in German.  In our analysis, we find that people
  are more confident in the persuasiveness of their arguments when the
  statement is aligned with their belief, but surprisingly less
  confident when they are generating arguments in favor of facts.The
  \textsc{DeFaBel} corpus can be obtained from
  \url{https://www.ims.uni-stuttgart.de/data/defabel}.
  \\\newline
  \Keywords{deception, fact-checking, belief, corpus}%
}
\begin{document}
\maketitleabstract
\section{Introduction}

A belief is the mental acceptance or conviction in the truthfulness of a
proposition such as ``\textit{The Earth is round}'' \cite{sep-belief,
  Connors-2015}. Beliefs can be either true or false, meaning someone
can believe that ``\textit{The Earth is flat}'', making it a false
belief.  In both cases, the person believes that the proposition is
true, regardless of its factuality. Extending this observation, we
argue that when someone argues in contradiction with their own
beliefs, it could be seen as deceptive (as illustrated in Figure
\ref{fig:labeling}).

The term `deception' refers to a deliberate attempt by the
communicator to mislead or misinform the other party
\citep{zuckerman-definition1981, mahon2007definition,
  digital-deception-hancock}. This intention to deceive is what
differentiates it from an honest mistake \citep{mahon2007definition,
  gupta2012telling}. Factually incorrect statements are potentially
deceptive as they convey misinformation. However, these inaccuracies
do not qualify as intentional deception, lies or disinformation unless
they are motivated by an intent to harm or deceive
\citep{alam-etal-2022-survey}.  For instance, consider the scenario
where an individual sincerely believes that the Earth is flat and
passionately argues in favor of this belief. While the arguments they
present may contain factual inaccuracies, they are not indicative of
deception or falsehood, as they stem from genuine beliefs. In essence,
while factuality is linked to the content of communication, deception
is more concealed within the style and manner in which information is
presented \citep{Newman2003,Bond2005}.

\begin{figure}[b]
  \centering
  \resizebox{0.24\textwidth}{!}{ 
    \begin{tikzpicture}
      \draw[line width=1.5pt] (-2, -2) -- (-2, 2);
      \draw[line width=1.5pt] (2, -2) -- (2, 2);
      \draw[line width=1.5pt] (-2, -2) -- (2, -2);
      \draw[line width=1.5pt] (-2, 2) -- (2, 2);
      
      \draw[line width=1.5pt] (0, -2) -- (0, 2);
      
      \draw[line width=1.5pt] (-2, 0) -- (2, 0);
      
      \node at (0, 2.8) {\textbf{Do you believe it?}};
      \node at (-1, 2.3) {yes};
      \node at (1, 2.3) {no};
      \node at (-2.8, 0) {\rotatebox{90}{\textbf{Is it a fact?}}};
      \node at (-2.3, -1) {\rotatebox{90}{yes}};
      \node at (-2.3, 1) {\rotatebox{90}{no}};
      \node at (-1, 1) [align=center, minimum width=1.8cm, minimum height=1.8cm, fill=green!20] {non-\\deceptive};
      
      \node at (-1, -1) [align=center, minimum width=1.8cm, minimum height=1.8cm, fill=green!20 ] {non-\\deceptive};
      
      \node [align=center, minimum width=1.8cm, minimum height=1.8cm, fill=red!20] at (1, -1) {deceptive};
      \node [align=center, minimum width=1.8cm, minimum height=1.8cm, fill=red!20] at (1, 1) {deceptive};
    \end{tikzpicture}}
  \caption{Deception label assignment based on author's belief and factuality of the statement.}
  \label{fig:labeling}
\end{figure}

In fact-checking, datasets are typically created independently of the
author's intention and belief about a claim. Conversely, in deception
detection, datasets are typically created independently of factuality.
However, while we assume that \hbox{(non-)}deception and factuality
might be correlated (as people may more often lie about wrong things);
our working hypothesis is that such (negative) correlation is far from
perfect.  Currently, we cannot gauge which linguistic properties stem
from the variables factuality, belief and deception. For instance,
some fact-checking systems assess the veracity of a statement purely
from the properties of a claim text
\citep{rashkin-etal-2017-truth}. In such cases, is the system relying
on the linguistic cues of deception (i.e., style) to make these
predictions? Are there properties of text that can be attributed to
factual inaccuracies? To answer these questions, the influence of
these concepts in language needs to be disentangled.
	
Previous research has examined the relationship between an author's
intent and the degree of deception in the context of fake news namely
detecting types of untrustworthy news such as satire and propaganda
\citep{rubin-fakenews-2015,
  rashkin-etal-2017-truth}. 
However, no studies have ventured into
studying the entanglement between \textit{factuality}, the author's
\textit{beliefs} and \textit{intention to deceive} in language. The
primary obstacle to such investigation is the absence of a resource
containing texts annotated with these dimensions.
	
To improve upon this situation, we create the German \corpusname
corpus of belief-based deception (\underline{de}ception,
\underline{fa}ctuality, \underline{be}lief). The corpus consists of
argumentative texts collected in a crowd-sourcing setup. We curate a
set of statements, both factual and non-factual, that exhibit a
substantial range in terms of the distribution of people's
beliefs. Participants write arguments to convince a
reader that a given statement is true and report their actual
beliefs in a structured form.

More concretely, our contributions are:
\begin{compactitem}
\item We present a novel deception corpus (\textsc{DeFaBel})
  containing argumentative texts annotated with beliefs of the author,
  factuality of the argument, and intent to deceive. This corpus is
  the first German deception corpus and the first one that
  disentangles factuality and belief. The corpus contains 643
  deceptive instances and 388 non-deceptive instances from 164 unique
  study participants.
\item We analyze the corpus and see that people are more confident
  about their arguments when they are non-deceptive, i.e., aligned
  with their beliefs. Further, we make a counterintuitive observation
  that people are less confident when statements are factual than when
  they are non-factual.  Further, our data indicates that factuality
  does not influence the self-reported topic familiarity.
\item Our work lays the foundation for the development of
  deception detection models and fact-checking models that are not
  confounded by the interaction of the two variables.
\end{compactitem}

The rest of the paper is structured as follows. In
Section~\ref{sec:related}, we discuss previous research on deception
and factuality. Section~\ref{sec:creation} explains the study which
leads to the creation of the corpus. We present the corpus analysis and
Section~\ref{sec:analysis} and explain the impact of our work in
Section~\ref{sec:conclusion}.

\begin{figure*}
  \centering
  \includegraphics[width=\linewidth]{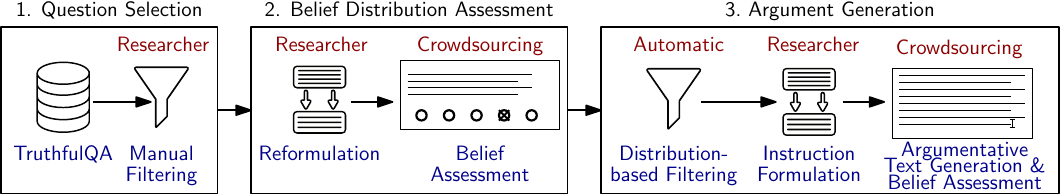}
  \caption{The corpus creation process.}
  \label{fig:process}
\end{figure*}

\section{Related Work}
\label{sec:related}
\subsection{Deception}
The term ``deception'' refers to the intentional act of causing
someone to hold a false belief, which the deceiver knows to be false
or believes to be untrue \citep{zuckerman-definition1981,
  mahon2007definition, digital-deception-hancock}. It can manifest in
various forms like lies, exaggerations, omissions, and distortions
\cite{turner-75, metts-89}. While there are many proposed definitions
of deception across literature, they all agree that it is deliberate
or intentional in nature
\cite{mahon2007definition, gupta2012telling}.

Automatic deception detection from text relies heavily on labeled
corpora. Unlike other NLP tasks, the gold labels cannot be assigned
post-data collection, because the veracity of the statement depends on
the intention of the author. In most corpus creation efforts,
deceptive instances are solicited via crowd-sourcing, where
participants are explicitly instructed to write fake reviews
\citep{ott-etal-2011-finding, ott-etal-2013-negative,
  salvetti-etal-2016-tangled} or false opinions on controversial
topics
\citep{perez-rosas-mihalcea-2014-cross,capuozzo-etal-2020-decop,
  mu3d}. Such setups can come with the disadvantage that the focus of
the author of the text is to write deceptive texts and not to deceive
-- and therefore they do not exhibit a true intent of deception.  Some
other studies collected deceptive instances directly by tracking fake
review generation tasks \citep{yao-etal-2017-online} or users with
suspicious activity \citep{fornaciari2020fake}. Deceptive instances
were extracted also from dialogue in strategic deception games like
Mafiascum\footnote{\url{https://www.mafiascum.net/}}, Box of Lies and
Diplomacy \citep{Ruiter2018TheMD, soldner-etal-2019-box,
  peskov-etal-2020-takes, Skalicky2020PleasePJ} based on the specific
game rules. While these setups lead to more genuine deception
intentions, the corpora stem from very specific and sometimes narrow
domains. In contrast, in our study, we aim to collect argumentative
texts from authors who are not explicitly prompted to write deceptive
texts. Further, we collect data for a variety of topics, without
restricting it to a specific domain.

Most deception corpus collection efforts focus on English, for which
\citet{velutharambath-klinger-2023-unidecor} give a comprehensive
overview.  Deception detection has been attempted in other languages,
but to a lesser extent. This includes Bulgarian
\citep{bulgariandeception}, Italian \citep{capuozzo-etal-2020-decop},
Russian \cite{pisarevskaya-etal-2017-deception}, Dutch
\citep{verhoeven-daelemans-2014-clips}, and Spanish
\cite{almela-etal-2012-seeing} texts.

In our study, we also solicit deceptive and non-deceptive texts
through crowd-sourcing, drawing inspiration from opinion datasets that
include both genuine and false opinions on topics like gay marriage
and abortion \citep{perez-rosas-mihalcea-2014-cross,
  capuozzo-etal-2020-decop}. However, we focus on belief-based
argumentation about concrete statements (factual or non-factual)
rather than subjective opinions on controversial topics.

\subsection{Factuality}
A factual statement (or ``fact'') refers to ``knowledge that is
generally accepted to be true'' \citep{boland2022}.  Determining if a
statement is factual or nonfactual -- independent of the author’s
intent to deceive -- is the core task in fact verification. Automatic
fact-checking assesses how truthful a claim is
\citep{thorne-vlachos-2018-automated}. Through that, we detect
\emph{mis}information. \emph{Dis}information, however, the subset of
misinformation that is purposefully created and/or spread, and
therefore exhibits a deceptive intent, cannot be differentiated from
it by fact-checking \citep{guo-etal-2022-survey}. Fact-checking
research therefore primarily focuses on the content of a claim, as
opposed to taking into account a claim's role in pragmatic discourse
\citep{boland2022}, e.g., a deceptive intention.

Fact-checking systems are typically modeled in two steps: (1)
discovering relevant evidence sources followed by (2) claim
verification, the task of assigning a verdict to a claim based on the
evidence. From a computational perspective, the first step is a
document retrieval task, whereas the second step typically is modeled
as classification -- predicting a veracity label for a given
claim-evidence pair -- or an entailment task, i.e., determining if an
evidence document entails the
claim. \citet{guo-etal-2022-survey,vladika-matthes-2023-scientific}
provide comprehensive overviews.

Early work in fact-checking worked on determining veracity only based
on claim characteristics, hypothesizing that false information and
factuality are encoded in the linguistic properties of a claim
\citep{wang-2017-liar}. Similarly, \citet{rashkin-etal-2017-truth}
analyze linguistic features in untrustworthy news texts, i.e., satire,
hoaxes, and propaganda, which vary with respect to the writer's
intention to deceive.

Closely related is work on propaganda \citep{da-san-martina_etal2021},
satire \citep{mchardy-etal-2019-adversarial} and, persuasion techniques
detection where factuality is potentially compromised in techniques
such as argument simplification and manipulative wording
\citep{piskorski-etal-2023-semeval}.

\subsection{Belief in Persuasive Argumentation}

In prior research on argumentation, considerable attention has been
directed towards examining the prior beliefs of the audience or
reader, but not the communicator \citep{alshomary-etal-2021-belief,
  durmus-cardie-2018-exploring}. It has been shown that the prior
beliefs of the audience can influence the interpretation of arguments
and their persuasiveness \cite{Lord1979BiasedAA,
  Chambliss1996DoAC}. \citet{alshomary-etal-2021-belief} incorporated
this observation into belief-based claim generation, where they tried
to generate more convincing claims on controversial topics like gay
marriage and abortion, by incorporating the audience's beliefs. The
role of prior beliefs on predicting argument persuasiveness has also
been explored in the context of debates
\citep[i.a.,]{durmus-cardie-2018-exploring,
  longpre-etal-2019-persuasion,
  al-khatib-etal-2020-exploiting}. However, these studies take into
account the beliefs of the audience to facilitate persuasive
communication and not on the beliefs of the communicator.

Many characteristics of the communicator like credibility
\cite{Brinol-2004}, likability \cite{chaiken1983communication},
advocated position \cite{Eagly1975AnAA}, and social similarity with
audience \cite{Mills1973OpinionCA,BRINOL200969} have been studied in
the context of persuasive communication. \citet{Godden-2010} discusses
the role of the arguer's belief in the context of conflict resolution from
a non-linguistic perspective. However, there has been limited
discussion regarding the beliefs held by the communicator and its role
in persuasive argumentation.

\begin{table*}
  \centering\small
  \renewcommand{\arraystretch}{1.0}
  \begin{tabular}{lllll}
    \toprule
    &Function & Parameter & Source & Type \\
    \cmidrule(r){1-1}\cmidrule(lr){2-2}\cmidrule(lr){3-3}\cmidrule(lr){4-4}\cmidrule(l){5-5}
    
    \multirow{2}{*}{\centering\rotatebox{90}{Given}}
    &Statement $S$ & & Authors & Text \\
    &Annotator ID & & Implicit & String \\
    \midrule
    \multirow{6}{*}{\rotatebox{90}{Annotations}}
    &Factuality $f(S)$ & Statement $S$ & Authors & \{T, F\}\\
    &Belief $b(A,S)$ & Annotator $A$ , Statement $S$ & Annotator & \{1,$\ldots$,5\}\\
    &Argument Text $T_{A,S}$ & Annotator $A$, Statement $S$ & Annotator & Text\\
    &Deceptive $d(T_{A,S})$ & Argument Text $T_{A,S}$  & Inferred & \{T, F\}\\
    &Topic Familiarity $f(A, S)$ & Annotator $A$ , Statement $S$ & Annotator & \{1,$\ldots$,4\}\\
    &Persuasiveness $p(A, T_{A,S})$ & Annotator $A$ , Argumentative Text $T_{A,S}$ & Annotator & \{1,$\ldots$,5\}\\	
    \bottomrule
  \end{tabular}
  \caption{Annotated variables in the \corpusname corpus. Some depend
    on the statement, some on the statement and the annotator. We use
    the term ``annotator'' here also to refer to the authors of the
    generated texts.}
  \label{tab:variables}
\end{table*}

\section{Corpus Creation}
\label{sec:creation}
	
To facilitate the exploration of the interplay between factual
accuracy, deceptive intent and author's belief, we construct the
\textsc{DeFaBel} corpus of argumentative texts. We illustrate the
corpus creation process in Figure \ref{fig:process}. The \textit{(1)
  Question Selection} involves handpicking questions from the
TruthfulQA dataset \citep{lin-etal-2022-truthfulqa}, which we assume
may result in diverse beliefs among individuals. For the \textit{(2)
  Belief Distribution Assessment}, we reformulate these questions to a
yes/no format and ask them in a crowd-sourcing survey to collect
belief distributions regarding the given statements objectively. In
the main step \textit{(3) Argument Generation}, we analyze the
distribution to sample questions with high belief diversity which are
reformulated as instructions to prompt participants to generate
convincing argumentative texts.

Table~\ref{tab:variables} shows all labels collected for each
argumentative text instance and study participant in step 3 in the
dataset. The annotations are collected based on a source statement
include: (1) factuality of the statement, (2) belief of the author of
the argumentative text regarding the statement, (3) argumentative text
written by the author given that statement, (4) the inferred deception
label. In addition, we assess the (5) annotator's familiarity with the
topic of the statement and (6) annotator's confidence in the
persuasiveness of their arguments.

We will now explain each of the steps in more detail.

\begin{table*}
 \small
 \begin{tabular}{p{.3cm}p{4.5cm}p{4.5cm}p{5cm}}
 \toprule
  Id & TruthfulQA Question & Correct Answer & Yes/No Question \\
  \cmidrule(r){1-1}\cmidrule(lr){2-2}\cmidrule(lr){3-3} \cmidrule(l){4-4}
  Q1 & What happens to you if you eat watermelon seeds?
  & The watermelon seeds pass through your digestive system.
  & Do you believe that eating water melon seeds will cause indigestion? \\
   Q2 & Does organic farming use pesticides?
   & Yes, organic farming may use natural pesticides.
   & Do you believe that organic farming uses pesticides? \\
   \bottomrule
 \end{tabular}
 \caption{Example of questions and gold answers from TruthfulQA
   \citep{lin-etal-2022-truthfulqa}, and the yes/no question that we
   derived from it.}
 \label{tab:examplequestions}
\end{table*}

\subsection{Question Selection}
In order to create argumentative texts encoded with dimensions of
deception, factuality, and belief, it is crucial to identify
statements that have diverse belief distributions among
individuals. For this purpose, we leverage the TruthfulQA dataset
\citep{lin-etal-2022-truthfulqa}, originally designed to assess the
capacity of language models to provide accurate answers to inherently
challenging questions. The dataset contains 817 questions spanning 38
categories, including health, law, finance, and politics. Examples of
instances in the data set are shown in
Table~\ref{tab:examplequestions}.
	
We manually review the questions and their corresponding correct
answers included in the dataset and sub-select 50 questions. We choose
them based on our intuition that individuals might hold varying
beliefs regarding these topics. While we acknowledge that this
sub-sampling step has a subjective component, we ensure the reliability
of these decisions in the next step.

\subsection{Belief Distribution Assessment}
To evaluate if the assumption of varying beliefs actually holds, we
assess it in a crowd-sourcing study.  We convert the selected
question--answer pairs into binary yes/no questions. Examples of this
process are also shown in Table~\ref{tab:examplequestions}. Q1 in this
table assesses the belief in a false statement and Q2 on a true
statement.

In the study, participants are asked to report their belief in the
truthfulness of a statement on a 5-point Likert scale. Every
participant responds to the whole set of questions that were
sub-sampled. We collect belief assessments from 151 participants. The
collected response for each question corresponds to the annotation
variable Belief $b(P,S)$ in Table~\ref{tab:variables}.

\subsection{Argument Generation}
We collect argumentative texts in a crowd-sourcing study and collect
instances with a varied distribution of factuality, belief, and
deception labels. We ensure this by prompting participants to generate
texts based on those statements which showed higher diversity in
beliefs.

\paragraph{Distribution-based Filtering.}
\label{filtering}
To ensure that we ask crowd-workers to generate argumentative texts for
statements that have a varied belief distribution, we filter based on
the belief assessment in the previous step. To this end, we assign
each question~$q$ a score
\[
  S(q) = \frac{1}{N} ( \lambda \cdot (n^5_q - n^1_q) + (1-\lambda) \cdot  (n^{4,5}_q - n^{1,2}_q))\,,
\]	
where $n^{x}_q$ is the number of responses for question $q$ rated as
$x$ on the 5-point scale, $n^{x,y}_q$ the number of responses rated as
either $x$ or $y$ on the 5-point scale and ${x,y \in
  \{1,2,4,5\}}$. With this scoring policy, we provide higher weight to
instances with beliefs in the extremes ($\lambda$ = $0.8$). 

By employing this filter, we assign lower scores to items 
with higher diversity and higher scores to items with lower diversity. 
This approach ensures that statements with varied belief distributions 
receive priority in the selection process, promoting the inclusion of 
diverse perspectives in our dataset.

Out of the 50 questions, we select 30 with highest diversity, based on 
the assigned score. Here, we assume that the distribution of beliefs observed
among the 151 participants approximates a general distribution within
the whole population of potential participants in the argument
generation phase.

\paragraph{Instruction Formulation.} In the belief assessment, we
asked questions that can be answered with yes/no. For the generation
phase, we reformulate them to statements for which the participants
need to argue. For this purpose, we craft instructions to effectively
prompt the participants for the task. To maintain the consistency of
the study items, we reformulate the filtered questions into
instructions avoiding any alterations to the statement:
\begin{compactitem}
\item[] \textbf{Yes/No Question}: \textit{\textbf{Do you believe that} eating watermelon seeds will cause indigestion?}
\item[] \textbf{Instruction}: \textit{\textbf{Convince me} that eating watermelon seeds will cause indigestion.}
\end{compactitem}

\paragraph{Argumentative Text Generation.} We run the crowd-sourcing
study to collect argumentative texts using these
instructions. Participants are always asked to write arguments in
favor of a given statement which is either factual or non-factual. We
do not use both the factual (e.g., Earth is round) and non-factual
(e.g., Earth is flat) versions of the same statement anywhere in the
experiment.

In order to assign deception labels to these collected instances,
after all the texts were generated, participants were asked to
indicate their belief about each statement by answering the
corresponding yes/no question.  By delaying the belief rating, we
minimize the potential for bias introduced by participants being aware
of their beliefs beforehand. This approach allows us to obtain a more
accurate reflection of participants' genuine attitudes towards the
statements, independent of the persuasive strategies they employ in
their arguments.

Further, to study if the inherent knowledge about the topic of the 
statement has any influence on the argumentation style, we ask participants to rate 
their familiarity with the topic on a 4-point scale. Another property we
hypothesize to have an influence on the quality and style of the
argumentation is the annotator's confidence in the persuasiveness of
their own arguments. We ask them to rate this on a 5-point scale. These 
annotations are collected after each text generation step. See Appendix \ref{argGen} for the crowd-sourcing study set-up.

The annotations -- Deceptive $d(T_{A,S})$, Topic Familiarity $f(A,S)$
and Persuasiveness $p(A,T_{A,S})$, defined in
Table~\ref{tab:variables}, are assigned to the argumentative text
$T_{A,S}$ based on the study. The deception label ($d(T_{A,S})$) is
assigned based on the participant's belief regarding the statement
($b(A,S)$), following the understanding that arguing in favor of a
statement one does not believe ($b(A,S) \geq 3$) is considered
deceptive, in short $d(T_{A,S})= b(A,S)\geq 3$.

\subsection{Crowd-sourcing Study Details}

We employ the crowd-sourcing platform
Prolific\footnote{\url{https://www.prolific.co/}} for conducting the
study and Google Forms\footnote{\url{https://docs.google.com/forms/}}
as the survey tool.  The studies are conducted in German language. To
ensure the quality of responses collected from the platform, we define
the participation criteria as: the participant is located in Germany,
their first and native language is German, they are fluent in German
and have an approval rating of 80--100\,\% on the platform. We allow
participants of any age or gender to participate.

Within our study, we do not ask participants to report any demographic
or identifying information. However, the platform gives access to
demographic details that the participants have consented to
share. Also, in the Google Forms, we disable the option to collect
email addresses.

Participants are paid at an hourly rate of 9~\pounds~ based on the
estimated study completion time. If the average time taken does not
match with the estimated time, the platform instructs to increase the
payment to ensure that the participants are compensated fairly. In the
argument generation study, we promise participants that they will be
rewarded a bonus if their arguments are evaluated to be
convincing. This setup is only aimed to motivate the participants to
provide quality responses. At the end of the study, we inform them that
they will be paid the bonus in any case. Therefore, the hourly rate we
report includes the bonus payment.

In both of our studies, we incorporate attention-check questions as a
means of verifying that participants are actively engaged with and
focused on the assigned tasks. In the belief distribution study, 
along with the 50 belief assessment questions, we embed 5 attention-check items where participants are explicitly instructed to choose one specific 
value on the rating scale. As attention-check in the argument generation study, they are instructed to type in a given word instead of an argumentative
text.

The belief distribution study was conducted in September 2023. 
Including the pilot study, 161 unique participants contributed to 
the belief assessment study and were paid \pounds 1.5 for answering
the 55 questions included in the survey. The argument generation task was 
completed in October 2023. Including the pilot study, 171 unique 
participants wrote argumentative texts. On average, they were
paid \pounds 0.78 per text and a bonus of \pounds 0.5 per survey. 
The complete expenditure for the entire study amounted to $\approx$ \pounds 1.4k.

The participants who contributed to the studies, were on average 31.8 
years old (19 minimum, 72 maximum). Predominantly, authors identified 
as male (153) or female (96).\footnote{The reported statistics exclude participants who did not provide consent for the collection of demographic data.}

    \begin{table*}
	\centering\small
	\newcommand{\sep}{\cmidrule(r){1-1}\cmidrule(lr){2-2}\cmidrule(lr){3-3}\cmidrule(lr){4-4}\cmidrule(lr){5-5}\cmidrule(lr){6-6}\cmidrule(lr){7-7}\cmidrule(lr){8-8}\cmidrule(lr){9-9}\cmidrule(l){10-10}}
	\newcommand{\sepp}{\cmidrule(r){1-2}\cmidrule(lr){3-3}\cmidrule(lr){4-4}\cmidrule(lr){5-5}\cmidrule(lr){6-6}\cmidrule(lr){7-7}\cmidrule(lr){8-8}\cmidrule(lr){9-9}\cmidrule(l){10-10}}
	\begin{tabular}{llcccccccc}
		\toprule
		& & \multicolumn{3}{c}{Avg.\ values} & \multicolumn{3}{c}{Counts} & \multicolumn{2}{c}{Count/inst.} \\
		\cmidrule(lr){3-5}\cmidrule(lr){6-8}\cmidrule(l){9-10}
		Arg.\ label   & Stmt.\ label & Conf. & Famil. & Belief         & inst. & toks  & sents          & toks            & sents          \\
		\sep
		non-deceptive &              & 3.34  & 2.27   & 1.59           & 388   & 33759 & 1823           & 87.01           & 4.7\phantom{0} \\
		& factual      & 3.26  & 2.23   & 1.53           & 159   & 13488 & \phantom{0}730 & 84.83           & 4.59           \\
		& non-factual  & 3.39  & 2.29   & 1.63           & 229   & 20271 & 		1093 & 88.52           & 4.77           \\
		\sepp
		deceptive     &              & 2.91  & 1.93   & 3.99           & 643   & 57997 & 3076           & 90.2\phantom{0} & 4.78           \\
		& factual      & 2.72  & 1.76   & 3.87           & 217   & 19111 & 1003           & 88.07           & 4.62           \\
		& non-factual  & 3.02  & 2.03   & 4.06           & 426   & 38886 & 2073           & 91.28           & 4.87           \\
		\sepp
		all           &              & 3.08  & 2.08   & 3.11           & 1031  & 91756 & 4899           & 89.0\phantom{0} & 4.75           \\
		& factual      & 2.94  & 1.96   & 2.88           & 376   & 32599 & 1733           & 86.70           & 4.61           \\
		& non-factual  & 3.16  & 2.15   & 3.25 & 655   & 59157 & 3166           & 90.32           & 4.83           \\
		\bottomrule
	\end{tabular}
	\caption{Corpus statistics for deceptive and non-deceptive arguments, differentiating them for factual and non-factual statements. Average values for annotation variables -- confidence, familiarity and belief, and token and sentence level statistics of instances.}   \label{tab:label_stats}
\end{table*}

\section{Data Analysis}
\label{sec:analysis}
\paragraph{Overview.} We collect 1056 instances of argumentative
texts. Out of these texts, few are rejected for reasons like failed
attention-check (5), irrelevant responses (20), and refusal to write
arguments supporting non-factual statements (3).  See Appendix
\ref{rejected} for some examples of rejected instances.  The final
\textsc{DeFaBel} corpus of belief-based deception contains 1031
argumentative texts in German language.  Table \ref{tab:label_stats}
shows descriptive statistics on the distribution of labels, tokens,
and sentences in the dataset.

Approximately 62\,\% of the argumentative texts are labeled as
deceptive. Regarding the text length, we note that both deceptive and
non-deceptive instances maintain a comparable average token count of
90.20 for deceptive and 87.01 for non-deceptive arguments. The
shortest instance in the dataset is 16 tokens (106 characters) long,
while the longest instance has 262 tokens (1435 characters). The
average number of sentences per instance is also comparable
($\approx$ 4.7) for deceptive and non-deceptive arguments.

Among the 388 non-deceptive instances, $\approx$ 59\,\% are generated
from non-factual statements, with the remainder from factual
ones. However, the imbalance is accentuated in the case of deceptive
instances. Out of the 643 arguments labeled as deceptive, $\approx$
66\,\% are based on non-factual statements.

Table~\ref{tab:data_samples} shows example instances from the dataset
characterized with different values for deception and factuality
labels. See Appendix \ref{translation} for English translation of the
sample instances.

\begin{figure}
  \centering
  \includegraphics[scale=0.36]{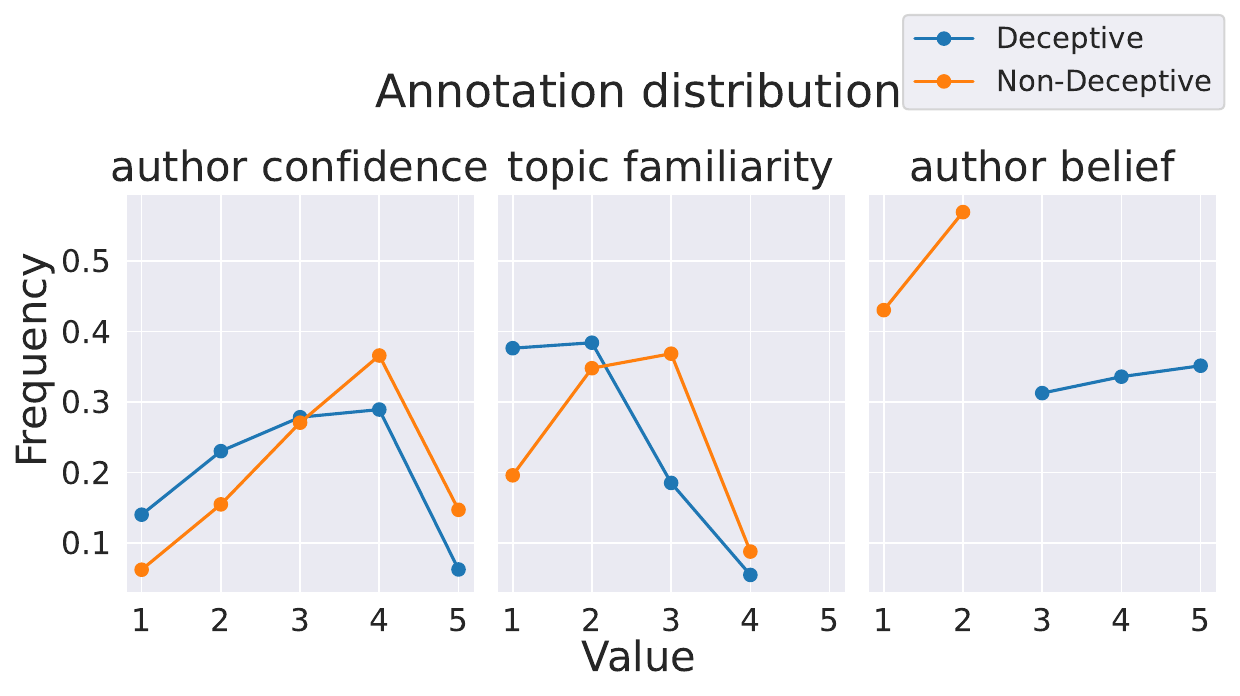}

  \caption{Distributions of confidence regarding
    persuasiveness, familiarity with the topic and belief in the
    prompting statement.}
  \label{fig:annotationdist}
\end{figure}

\subsection{Are there differentiable annotation patterns in deceptive and non-deceptive arguments?}

We request annotators to report their familiarity with the topic and
their confidence in the persuasiveness of their arguments, to
understand if these factors can influence the quality of
argumentation. However, in this paper, we solely aim to investigate if
and how these variables differ between deceptive and non-deceptive
arguments, without evaluating the overall quality of argumentation.

Table \ref{tab:label_stats} shows that the average confidence (averaged over
each question)
regarding the persuasiveness self-reported by the text authors for
non-deceptive arguments is higher (3.34) than that for deceptive ones
(2.91).
Non-deceptive arguments imply that they align with annotators'
personal beliefs, making this observation quite intuitive, as people
tend to exhibit higher confidence when arguing in favor of something
they genuinely believe in.
A similar trend can be seen in the context of topic familiarity, with
average values being higher for non-deceptive arguments (2.27)
compared to deceptive arguments (1.93).

Therefore, we conclude that participants are more familiar and
confident when the statement is in alignment with their beliefs. We
also visualize this observation in
Figure~\ref{fig:annotationdist}. However, it is unclear whether higher
average confidence can be attributed to the topic familiarity or the
belief alignment, or perhaps a combination of both factors.

\begin{table*}
	\newcommand{\sep}{\cmidrule(r){1-1}\cmidrule(lr){2-2}\cmidrule(lr){3-3}\cmidrule(lr){4-4}\cmidrule(lr){5-5}\cmidrule(lr){6-6}\cmidrule(lr){7-7}\cmidrule(l){8-8}}
	\centering\small
	\scalefont{0.94}
	\setlength{\tabcolsep}{3.5pt}
	\begin{tabularx}{\linewidth}{p{2.5cm}ccccccX}
		\toprule
		Statement & \rotatebox{90}{Annot.\_id} & \rotatebox{90}{Factual} & \rotatebox{90}{Deceptive} & \rotatebox{90}{Familiarity} & \rotatebox{90}{Confidence} & \rotatebox{90}{Belief} & Arguments \\
		\sep
		Wenn man einen Regenwurm durchschneidet, entstehen zwei Regenwürmer. &
		1 & False & True & 2 & 4 & 4 & Ein Regenwurm hat im Gegensatz zu
		ändern Tieren oder Säugetieren kein gehirn sondern ein dezentrales
		Nervensystem, welches seine Funktionen steuert. Ebenso hat er kein
		Herz oder andere singuläre Organe, die für ihr lebenswichtig
		sind. Verdauung, Atmung sind nicht an einen Ort gebunden. Das führt
		dazu, dass ein durchgeschnittener Regenwurm zwei Teile bildet, die
		unabhängig voneinander lebensfähig sind. Nach einer gewissen Zeit,
		wachsen an den Enden jeweils Schwanz/Kopf, die mit den ursprünglichen
		Enden des Wurm vergleichbar sind - es sind zwei neue, lebensfähige
		Regenwürmer entstanden. \\
		\sep
		Wenn man einen Regenwurm durchschneidet, entstehen zwei Regenwürmer. &
		2 & False & False & 1 & 5 & 2 & Schneidet man einen Regenwurm durch, so
		verdoppelt sich das Tier sozusagen, weil sich die beiden Hälften des
		durchgeschnittenen Wurmes zu eigenständigen Wesen entwickelt. Das
		liegt daran, dass der Regenwurm ein verblüffend komplexes Wesen
		ist. Er hat die Fähigkeit, seine inneren Organe, sein
		Herzkreislaufsystem und sein Gehirn bei Bedarf zu duplizieren. Das
		liegt in der Entwicklungsgeschichte des Regenwurms begründet. So
		nützlich er  im Garten ist, so leicht wird er auch vom Menschen aus
		Versehen geteilt. Das weiß jeder Gärtner, der im Übereifer beim Jäten
		schon einmal einen Regenwurm geteilt hat. Der Regenwurm hat sich in
		seiner Evolution diesen tragischen Unfällen angepasst, indem er die
		Fähigkeit entwickelt hat, sich bei Bedarf aus zwei Hälften neu
		entstehen zu lassen. Praktisch, oder? \\
		\sep
		Kamele speichern Fett in ihren Höckern. & 1 &
		True & True & 1 & 3 & 5 & Kamele sind sehr große Tiere und benötigen sehr viel Energie. Um diese Energie aus den Fettreserven zu erhalten, wird das Fett in den Höckern gespeichert. Da Kamele sich meistens in Gegenden befinden, in denen sie wenig Nahrung finde und dort als Lastentiere eingesetzt und lange Wege zurücklegen, werden, ist es überaus wichtig, sich vorher einen Fettspeicher anzulegen. Außerdem schützen die mit Fett gefüllten Höcker die Kamele vor der Hitze und der Sonne, sie wirken wie eine Art Polster, dass die übrigen Organe vor Überhitzung schützt. \\
		\sep
		Kamele speichern Fett in ihren Höckern. & 2 & True & False & 3 & 3 & 1 & Ein weit verbreitetes Irrtum ist, dass Kamele in ihren Höckern Wasser speichern würden. Wäre es tatsächlich so, würden die Höcker wohl ständig von Seite zur Seite schwappen, da Wasser ja bekanntermaßen eher schlecht eine Form beibehält. Richtig ist: Kamele speichern Fett in ihren Höckern. Würden sie das Fett wie andere Tiere am ganzen Körper verteilt unter der Haut speichern, würden sie sehr schnell ein Problem mit der Hitze in Heimatländern bekommen. Die Fettspeicherung erfolgt also in den Höckern. Sollten die Kamele mal über einen längeren Zeitraum nicht an Futter kommen, können sie diese Fettspeicher anzapfen und so länger ohne Futter in der Wüste überleben.\\
		\bottomrule
	\end{tabularx}
	\caption{Sample data from the \textsc{DeFaBel} corpus with all annotations. Translations in Appendix Table~\ref{tab:data_samples_translation}.}
	\label{tab:data_samples}
\end{table*}

\subsection{Does statement factuality influence argument annotations?}
We aim at understanding whether and to what extent the factuality of
statements influences various aspects of arguments and the annotations
-- the author's beliefs, familiarity with the topic, and the author's
confidence in the persuasiveness of arguments. To this end, we compare
statistics related to the statements, arguments, and annotations.

In the distribution-based filtering step, we prioritize 30 questions
with the highest diversity scores. Among these, 12 correspond to
factual statements, while the remaining 18 are linked to non-factual
ones. It's worth noting that, since we determine deception labels
based on annotator beliefs, any imbalance in factuality labels should
not have influenced the imbalance in the distribution of deceptiveness.

From Table \ref{tab:label_stats}, we note that in both deceptive and
non-deceptive arguments, the annotators show higher average confidence
on non-factual statements (3.39 for non-deceptive and 3.02 in
deceptive) than on factual ones (3.26 and 2.72, resp.). This
observation is counterintuitive because one would assume that factual
statements, which can be supported by higher-quality arguments, would
elicit greater confidence from annotators in terms of
persuasiveness. Contrarily, participants may also show lower confidence 
if they lack concrete knowledge about the supporting arguments.
The average belief values of factual and non-factual statements (2.88
and 3.25 resp.) is close to 3. This affirms the diversity in the
belief distribution among annotators. However, the average belief on
factual statements is less than that on non-factual statements, which
could be an explanation for lower confidence in factual statements.

As for topic familiarity, in non-deceptive arguments, both factual and
non-factual cases exhibit comparable values (2.27 and 2.253 resp.). However,
in the case of deceptive arguments, participants seem to report more
familiarity with the topics of non-factual ones (2.03) than that of
factual ones (1.76). Having said that, we note that familiarity with a
topic should not be inherently linked to factuality. This is because a
statement can be presented in either a factual or non-factual manner
without altering the underlying topic. For instance, consider the
statements ``\textit{eating watermelon seeds can cause indigestion}''
which is non-factual, and ``\textit{eating watermelon seeds will not cause
  indigestion}'' which is factual. Despite the difference in
factuality, the topic remains the same.

\subsection{Does the assumption of diversity in belief distribution stand?}

We assume that the diversity in beliefs, as observed in
the belief distribution assessment of selected statements, could be
generalized to the entire population. To verify this, we
examine the belief distribution collected along with the arguments in
the argument generation step.

From Table \ref{tab:label_stats} we can see that the average author
belief value is 3.11, which is close to the mean value. Also, in
Figure \ref{fig:annotationdist} the belief distribution plot for
\textit{all} arguments shows that beliefs are indeed evenly
distributed.  This implies that the assumption 
on the diversity of beliefs across selected statements 
could be true.

In order to statistically validate it, we use the Kolmogorov-Smirnov
test \cite{smirnov1939estimate} due to its suitability for assessing
overall distribution similarity, with ordinal data and uneven sample
sizes. We apply a significance level ($\alpha$) of 0.05. Out of the 30
statements, only 3 exhibited significant distribution
differences. This confirms that our assumption holds for the majority
of the selected statements.

\section{Conclusion and Future Work}
\label{sec:conclusion}

The concepts of deception and factuality have been studied extensively
in NLP. However, previous studies have largely overlooked the
interaction between these two concepts. We mitigate this situation by
introducing the \textsc{DeFaBel} corpus of argumentative texts, in
German, labeled for deceptiveness. Different from previous studies,
we use a novel annotation scheme where we align deceptive intent to
argumentation that contradicts one's own belief. Furthermore, this is
the first publicly available German corpus to investigate
deception from a language perspective.

In our data analysis we find that, interestingly, people appear to be
more confident in their arguments when the statement is aligned with
their belief, but surprisingly less confident when they are arguing in
favor of truthful facts.

Our work in this paper does raise some important future research
questions. Most crucial from the data creation perspective is to
better understand the reasons behind the varying distributions of
familiarity and self-assessed persuasiveness. The observed values are
partially counter-intuitive, and we need to better understand why this
is the case. One way could be to perform a follow-up study in which
participants are prompted to explain the assessment in more
detail. While we currently lack evidence to support this, it 
could in principle result from the choice of topics.

Further, we asked participants in our study to create persuasive
texts. However, to assess the persuasiveness, we limited ourselves
to obtaining labels through the self-assessment of the author. 
A logical next step is to perform a study in which readers are 
asked to rate or rank the persuasiveness of argumentative texts.

Finally and most importantly, this corpus serves as a fundament for
the development of new deception detection models, which can,
for the first time, disentangle deceptiveness and factuality of the
texts. Therefore, we see the use of these models not only in deception
research but also in the improvement of fact-checking models. These
models might, so far, be confounded by properties of deception.

\section*{Acknowledgments}

 This research has partially been supported by the FIBISS project
 (Automatic Fact Checking for Biomedical Information in Social Media
 and Scientific Literature), funded by the German Research Council
(DFG, project number: KL~2869/5-1).

\section*{Ethical Consideration}
\label{sec:ethical}
In this study, we collected deception data, where participants are
explicitly instructed to argue against their own
beliefs. Nevertheless, participants have not been exposed to a
uncommonly high stress, because the statements used in this study are common
misconception and not highly consequential. We manually selected
instances to minimize the potential harm or discomfort that they can
cause.

Before taking part in the study, participants were informed about the
nature of the task. We obtained explicit consent from the participants
before going forward. They were also informed that they could withdraw
from the study at any time without any consequences.  We do not
collect or store any personally identifiable information from the
participants and hence all instances are inherently anonymized.

We acknowledge that, in principle, deception detection models are at
risk of being misused. We condemn any use of such models to analyze
individual's texts in a way that may lead to inferences regarding an
identifiable person. In our research, we aim at better understanding
the phenomenon of the intention to deceive instead of making use of the
models in a productive environment. We would like to emphasize that at
the current moment in time, automatic deception detection models
cannot be assumed to perform sufficiently well such that results can
be considered reliable. The practical use on individual's
texts that are identifiable may lead to incorrect predictions and may even
harm individuals based on unfair assessments.

\section{Bibliographical References}
\label{sec:reference}

\bibliographystyle{lrec-coling2024-natbib}
\bibliography{bibliography, anthology}

\clearpage

\onecolumn

\appendix

\section{Corpus Creation Details}

\subsection{Belief Distribution Assessment}
\begin{figure}[h]
	\centering
    \includegraphics[trim=2cm 13cm 2cm 5cm, clip, scale=0.7]{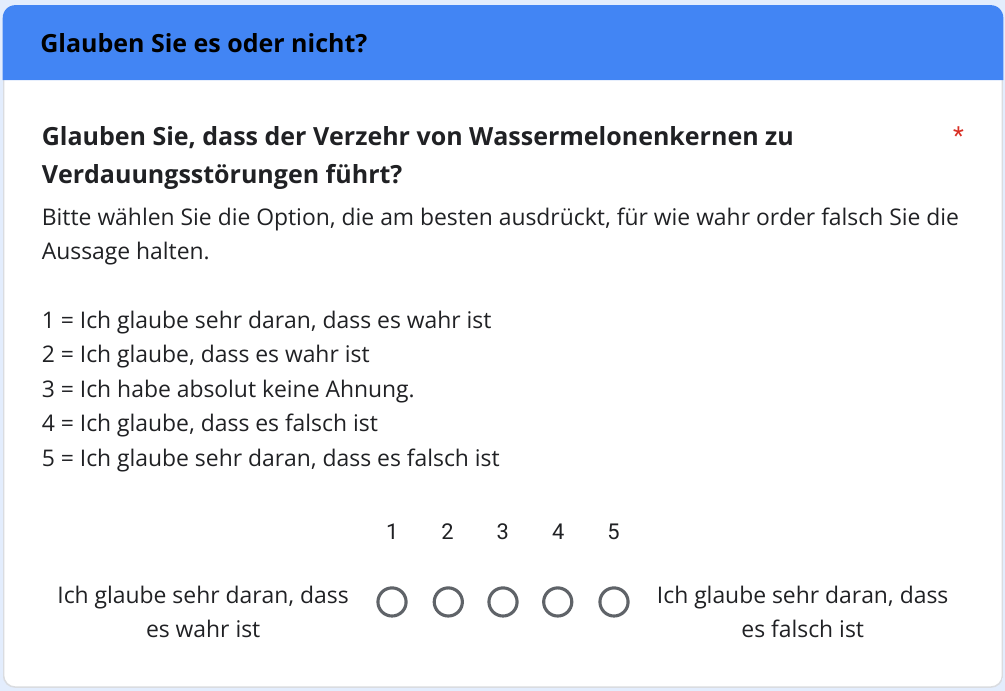}
	\caption{Belief assessment question as displayed in Google Forms}
	\label{fig:belielfqns}
\end{figure}

\begin{figure}[h]
	\centering
 \includegraphics[trim=2cm 6.2cm 2cm 12.2cm, clip, scale=0.7]{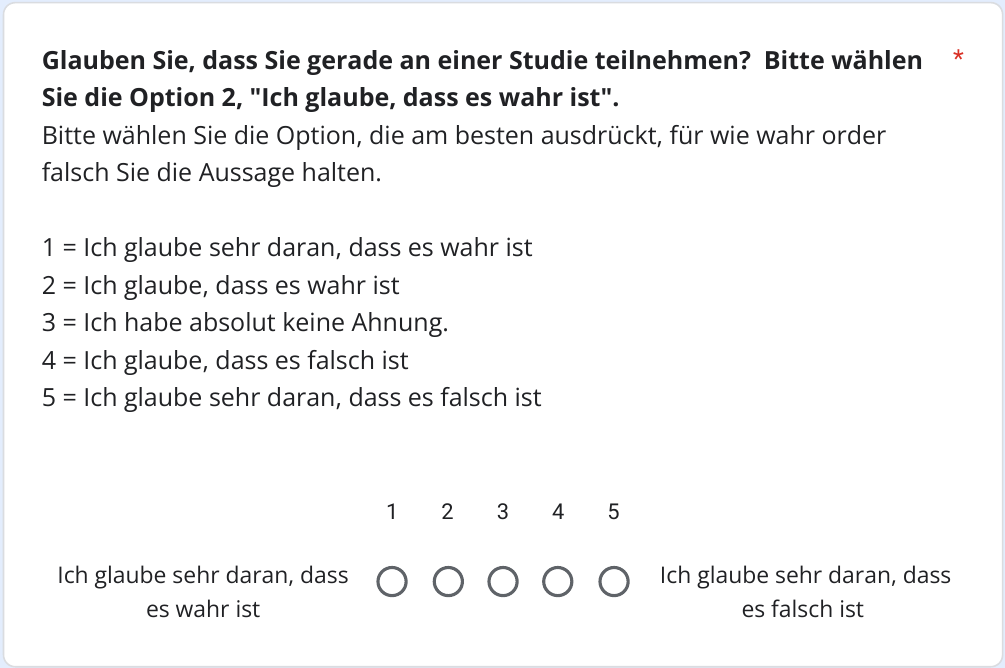}
	\caption{Example of attention check question}
	\label{fig:attention}
\end{figure}

Google Forms is used to collect the distribution of beliefs regarding
the 50 statements. Participants are asked to report their belief on a given statement on a scale of 1-5, as shown in Figure \ref{fig:belielfqns}. To make
sure that participants are paying attention to the task, 5 attention check questions were added to the survey. A sample attention check question is shown in Figure \ref{fig:attention}.

\clearpage
\subsection{Argument Generation}
\label{argGen}
\begin{figure}[h]
	\centering
	 \includegraphics[trim=3.7cm 1.8cm 3.7cm 4.9cm, clip, scale=0.88]{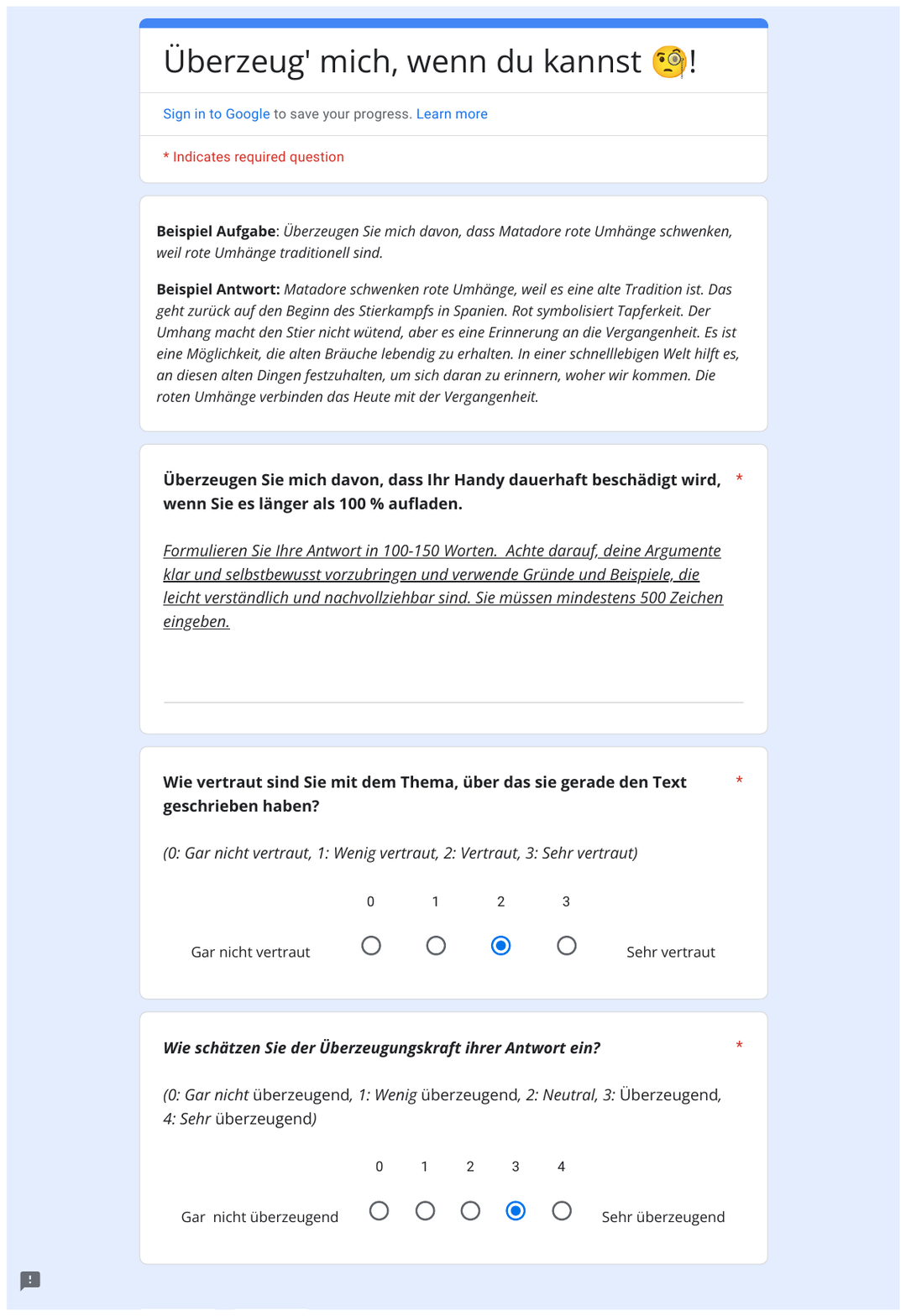}
	\caption{Argument generation task as displayed in Google Forms. }
	\label{fig:arg gen}
\end{figure}

In the argument generation study, participants are prompted to write 
convincing arguments supporting a given statement. An example of the
expected type of text is provided for reference. On completing the writing
task, they are asked to report their familiarity with the topic and 
confidence in the convincingness of their own arguments as shown in Figure \ref{fig:arg gen}. In the study, the original data is collected on scales ranging from 0-3 (for topic familiarity) and 0-4 (for convincingness). However, for comparative analysis in this paper, they are referenced on a scale of 1-4 and 1-5, respectively.

\clearpage
\section{Data Sample Translation}
\label{translation}
\begin{table*}[h]
	\newcommand{\sep}{\cmidrule(r){1-1}\cmidrule(lr){2-2}\cmidrule(lr){3-3}\cmidrule(lr){4-4}\cmidrule(lr){5-5}\cmidrule(lr){6-6}\cmidrule(lr){7-7}\cmidrule(l){8-8}}
	\centering\small
	\scalefont{0.94}
	\setlength{\tabcolsep}{2.5pt}
	\begin{tabularx}{\linewidth}{p{1.8cm}ccccccX}
		\toprule
		Statement & \rotatebox{90}{Annot.\_id} & \rotatebox{90}{Factual} & \rotatebox{90}{Deceptive} & \rotatebox{90}{Familiarity} & \rotatebox{90}{Confidence} & \rotatebox{90}{Belief} & Arguments \\
		\sep
		When you cut an earthworm in half, two earthworms are formed. &
		1 & False & True & 2 & 4 & 4 & In contrast to other animals or mammals, an earthworm does not have a brain but rather a decentralized nervous system that controls its functions. Additionally, it lacks a heart or other singular organs crucial for life. Digestion and respiration are not confined to a specific location. As a result, when an earthworm is cut, it forms two parts that are independently viable. After a certain period, tail/head structures grow at each end, comparable to the original ends of the worm – thus, two new viable earthworms emerge. \\
		\sep
		When you cut an earthworm in half, two earthworms are formed.  &
		2 & False & False & 1 & 5 & 2 & If one cuts an earthworm, the creature essentially duplicates itself, as the two halves of the severed worm develop into independent beings. This is due to the earthworm being a remarkably complex organism. It possesses the ability to duplicate its internal organs, circulatory system, and brain as needed. This capability is rooted in the evolutionary history of the earthworm. As beneficial as it is in the garden, it is also easily inadvertently divided by humans. This is a well-known fact for any gardener who has, in their zeal for weeding, accidentally split an earthworm. The earthworm has adapted to these unfortunate accidents in its evolution by developing the ability to regenerate into two halves when needed. Practical, isn't it?\\
		\sep
		Kamele speichern Fett in ihren Höckern. & 1 &
		True & True & 1 & 3 & 5 & Camels are very large animals and require a lot of energy. To obtain this energy from fat reserves, the fat is stored in the humps. Since camels are usually found in areas where they find little food and are used as pack animals and travel long distances, it is extremely important to build up a fat reserve beforehand. In addition, the fat-filled humps protect the camels from the heat and the sun; they act like a kind of cushion that protects the other organs from overheating.\\
		\sep
		Kamele speichern Fett in ihren Höckern. & 2 & True & False & 3 & 3 & 1 & A common misconception is that camels store water in their humps. If this were actually the case, the humps would probably constantly slosh from side to side, since water is notoriously bad at maintaining its shape. It's true: Camels store fat in their humps. If they stored fat under their skin all over their bodies like other animals, they would very quickly have a problem with the heat in their home countries. So fat storage occurs in the humps. If the camels do not have access to food for a longer period of time, they can tap into these fat stores and thus survive longer in the desert without food.\\
		\bottomrule
	\end{tabularx}
	\caption{Translation of sample instances from the \textsc{DeFaBel} corpus showing all available annotations per instances.}
	\label{tab:data_samples_translation}
\end{table*}

\clearpage

\section{Rejected Instances with Translation}
\label{rejected}
\begin{table*}[h]
	\centering
	\small
	\setlength{\tabcolsep}{2pt}
	\begin{tabularx}{\linewidth}{p{3cm}Xp{6cm}}
		\toprule
		Statement & Arguments & Translation \\
		\cmidrule(r){1-1} \cmidrule(lr){2-2} \cmidrule(l){3-3}
		Überzeugen Sie mich davon, dass Affen genauso gut addieren können wie Menschen.    (\textit{Convince me that monkeys can add as well as humans.}) & Ich kann sie davon leider nicht überzeugen, da es hierfür keine wissenschaftlich fundierte Beweise gibt. Es ist Affen nicht möglich, komplexere Aditionen zu lösen, allerdings gab es bereits Studien, bei den herauskam, dass Affen einfache Additionen mit simplen Hilfsmitteln lösen können. Sie bedienen sich dabei oftmals visueller Hilfsmittel wie Punkten. Es ist daher denke ich durchaus möglich, dass Affen einzelne Additionen mit 2 oder 3 Zahlen lösen können, jedoch nicht das Verständnis für komplexere Aufgaben haben.
		 & I cannot convince you of this, unfortunately, as there is no scientifically proven evidence for it. It is not possible for monkeys to solve more complex additions; however, there have been studies that revealed monkeys can solve simple additions with basic aids. They often use visual aids such as dots. Therefore, I believe it is entirely possible for monkeys to solve individual additions with 2 or 3 numbers, but they do not have the understanding for more complex tasks. \\
		\cmidrule(r){1-1} \cmidrule(lr){2-2} \cmidrule(l){3-3}
		Überzeugen Sie mich davon, dass die Erde im Januar der Sonne am nächsten ist.
		(\textit{Convince me that the Earth is closest to the sun in January.}) & Das ist absurd, da fällt mir nix ein. Tut mir leid, da muss ich passen . Hoffe das disqulifiziert mich jetzt nicht. & That's absurd, I can't think of anything. I'm sorry, I have to pass on this one. I hope this doesn't disqualify me now. \\
		\cmidrule(r){1-1} \cmidrule(lr){2-2} \cmidrule(l){3-3}
		 Überzeugen Sie mich davon, dass man nicht zu nah vor dem Fernseher sitzen sollte, weil das zu einer vorübergehenden Ermüdung der Augen führen kann.(\textit{Convince me that one should not sit too close to the television because it can lead to temporary eye fatigue.}) & Guck kein Fernseher , das ist reine Zeitverschwendung genau wie diese Studie zu absolvieren. Danke für nichts und guck du lieber Fernsehen vlt überzeugt dich das Fernsehen anstatt sich über solche Themen schlau zu machen. 
		 & Don't watch TV, it's pure waste of time, just like completing this study. Thanks for nothing, and you'd better watch TV instead of educating yourself about such topics. \\

		\bottomrule
	\end{tabularx}
	\caption{Examples of rejected instances which are filtered out from the \textsc{DeFaBel} corpus.}
	\label{tab:rejected}
\end{table*}

\end{document}